\documentclass{article}
\usepackage{arxiv}
\usepackage{graphicx}
\usepackage[utf8]{inputenc} 
\usepackage[T1]{fontenc}    
\usepackage{hyperref}       
\usepackage{url}            
\usepackage{booktabs}       
\usepackage{amsfonts}       
\usepackage{nicefrac}       
\usepackage{microtype}      

\title{Artificial Neural Networks and Adaptive Neuro-fuzzy Models for Prediction of Remaining Useful Life}

\author{
  Razieh Tavakoli\\
  Department of Civil Engineering \\
  The University of Texas at Arlington \\
  Box 19308, Arlington, TX 76019 \\
  \texttt{razieh.tavakoli@uta.edu} \\
  \And
 Mohammad Najafi \\
  Department of Civil Engineering \\
  The University of Texas at Arlington \\
  Box 19308, Arlington, TX 76019 \\
  \texttt{najafi@uta.edu} \\
 \AND
  Ali Sharifara \\
  Department of Computer Science and Engineering \\
  The University of Texas at Arlington \\
  Box 19308, Arlington, TX 76019 \\
  \texttt{ali.sharifara@uta.edu} \\
}

\begin{document}
\maketitle

\begin{abstract}
The U.S. water distribution system contains thousands of miles of pipes constructed from different materials, and of various sizes, and age. These pipes suffer from physical, environmental, structural and operational stresses, causing deterioration which eventually leads to their failure. Pipe deterioration results in increased break rates, reduced hydraulic capacity, and detrimental impacts on water quality. Therefore, it is crucial to use accurate models to forecast deterioration rates along with estimating the remaining useful life of the pipes to implement essential interference plans in order to prevent catastrophic failures. This paper discusses a computational model that forecasts the RUL of water pipes by applying Artificial Neural Networks (ANNs) as well as Adaptive Neural Fuzzy Inference System (ANFIS). These models are trained and tested acquired field data to identify the significant parameters that impact the prediction of RUL. It is concluded that, on average, with approximately 10\% of wall thickness loss in existing cast iron, ductile iron, asbestos-cement, and steel water pipes, the reduction of the remaining useful life is approximately 50\%

\end{abstract}

\keywords{Remaining useful life prediction \and  Artificial Neural Networks (ANNs) \and Adaptive Neural Fuzzy Inference System (ANFIS)}

\section{Introduction}
Pipe deterioration is an unavoidable process that occurs over time due to structural, operational and hydraulic capacity failure. Structural failure is caused by any kind of defects on the pipe wall that reduces the structural integrity of the pipe segment. Likewise, the soil surrounding the pipe has an essential role in failure time of pipes. In general, cracks, internal and external corrosion, pipe deflection, misaligned joints, and breaks are the most common type of defects associated with structural failure \cite{EPA2009}. Several factors impact the structural deterioration of water mains and their failures, including pipe material, pipe size, pipe age, soil type, climate, and cyclic pressures. However, the physical processes that cause pipe breakage are complicated and as most water pipes are buried, only very few research is available about how they deteriorate and fail \cite{Grigg2013}. Operational failure is the most common failure in a water distribution system and generally occurs by a physical cause and can be resolved during a maintenance procedure and usually does not affect the structural integrity of the pipes \cite{EPA2009}.

Statistical models have been developed to quantify the structural deterioration of water distribution pipes based on analyzing various levels of historical data (Shahata and Zayed, 2012). Rajani and Kleiner (2001) provided a critical review of the statistical, deterministic and probabilistic models that attempt to quantify and predict water pipe breakage or structural pipe failures. Their results illustrated statistical methods using historical data on water main breakage \cite{Kleiner2001}. Moreover, Martins \emph{et al.}, 2013; Osman \emph{et al.}, 2011, Tscheikner \emph{et al.}, 2011, compare the strengths, weaknesses, and limitations of those statistical models \cite{Martins2013,Osman2011,Tscheikner2016}. Most of the models use different strategies to handle scarce data situations, so even for limited data availability deterioration models can give valuable information. \cite{Winkler2018,Scheidegger2015, Scholten2013}. Other modeling categories are artificial intelligence models (e.g. genetic algorithms \cite{Nicklow2010}, neural networks \cite{Tran2007} or Neuro-fuzzy systems \cite{Christodoulou2010}. These are purely data-driven approaches that enable solving of complex problems without the necessity of detailed explicitly known model assumptions \cite{Winkler2018}.

Likewise, Kulandaivel (2004) developed a sanitary sewer pipe condition prediction model based on Neural Networks, which can prioritize sewer pipes for risk of degradation. Furthermore, he evaluated the performance of the proposed model with different sewer pipes data sources. The results showed that the prioritization model can assist municipal agencies in efficient utilization for targeting critical areas with predicting deteriorated sewer pipes \cite{mastersthesis}.

Fahmy and Moslehi (2009) proposed a model to forecast the remaining useful life of cast iron water mains. They considered several factors related to pipe properties, pipe operating conditions and the external environmental factors influencing pipe failures. Three different data-driven techniques were considered in their model development; these techniques are multiple regression and two types of artificial neural networks: multi-layer perception and general regression neural networks. The data was used in their model development acquired from 16 municipalities in Canada and the United States. They showed data-driven modeling methods are effective in forecasting the remaining useful life of cast iron water mains and it overcame limitations associated with existing models \cite{fahmy2009}.

Fares and Zayed (2010) developed a risk model for water main failure, which evaluated the risk associated with each pipe in a water network. They considered four main factors: environmental, physical, operational factors, and consequences of failure and 16 sub-factors. In order to develop the risk failure model, Hierarchical Fuzzy Expert System (HFES) was used to process the input data and generated the risk of failure index of each water main \cite{Fares2010}. They presented that pipe age was the most significant factor of water main failure risk, followed by pipe material and breakage rate. The model validation was 74.8\%, which was reasonable considering the uncertainty involved in the collected data \cite{Fares2010}.

Rogers (2011), proposed a performance-based approach to evaluate the present and future conditions of pipes from routine pipe inventory and break record data. The Multi-criteria Decision Analysis (MCDA) was performed to provide a pipe replacement ranking. He showed that the pipe failure prediction and MCDA modules are a quantifiable process for prioritizing short-term and long-term pipe renewal decisions \cite{Rogers2011}.

 Christodoulou and Deligianni (2009) presented a neuro-fuzzy decision support system for the performance of multi-factored risk-of-failure analysis and pipe asset management, as applied to urban water distribution networks. They showed that the combination of Neural Networks and Fuzzy logic is extremely effective in risk-of-failure analysis and preventive maintenance of water distribution networks \cite{Christodoulou2010}.
 
 Clair and Sinha (2011) presented a state-of-the-technology review on water pipe condition, deterioration, and failure rate prediction models to identify the gap between the existing models in previous research and those currently used by water agencies. They proved limitations of model capabilities and complexity in analysis and validating these models \cite{Clair2012}.

Osman and Bainbridge (2011), presented a comparison and analysis of rate-of-failure models (ROF) and transition-state models using a single dataset for cast-iron and ductile-iron pipes in the City of Hamilton, Ontario, Canada. They compared the models’ ability to support breakage forecasting, long-term strategic planning, and short-term tactical planning \cite{Osman2011}.

In above research, various variables are considered in the analysis, including pipeline diameter, material type, installation year, break history, etc., along with other risk factors such as operating pressure, soil type and soil acidity. However, there is no standard procedure for recording data on leaks, breaks, and condition indicators. Moreover, the above research does not consider wall thickness loss to predict the remaining useful life of water pipes.  According to previous studies about failure in water pipes and remaining useful life prediction, there is a lack of a comprehensive model to evaluate the risk of water pipes failure and determine the remaining useful life for different types of water pipe and there is no research using a combination of Artificial Neural Network (ANN) and Adaptive Neural Fuzzy Inference System (ANFIS) models to predict the remaining useful life of water pipes. In order to overcome the limitations of existing approaches, this paper aims to implement a new approach for prediction of remaining useful life of water pipes.

in this research, the Remaining Useful Life (RUL) is defined as an estimation before pipe experiences a structural failure. This type of failure involves conditions where the pipe section cannot function properly without replacement such as a pipe break. This research determines the most significant variables influencing RUL. Several variables are considered for the model development including pipe age, length, wall thickness loss, installation year, the number of breaks, and pipe materials for cast iron, ductile iron, asbestos cement and steel pipes with limited diameter ranges 4 in. to 24 in (100 mm to 635 mm).

\section{Data Analysis}
Preliminary data analysis is carried out; which consists of sorting and ordering of data obtained from sources, comparing data and finalizing input variables. The validation data are embedded into most appropriate regression models for comparing results with the actual results. 

Data for this research was collected from several municipalities. The first dataset was obtained from Southgate Water District (SWD) in Denver, Colorado. The SWD water system contained approximately 40\% AC (Asbestos Cement) pipe, 34\% DI (Ductile iron) pipe, 25\% PVC (Poly Vinyl Chloride) pipe and about 1\% of CI (Cast Iron) and steel pipe combined. These pipes had extreme wall thickness losses due to age and deterioration process. The second dataset was collected from a municipality in Laval, Moncton and Quebec, Canada, with deteriorated distribution pipes over a large area \cite{Wang2006}. Iron pipes that were installed in 1954 have 95 breaks from 1987 to 2001 and 11 breaks in 1999 all the breaks occurred when the pipes are of ages between 20 and 50 years \cite{Wang2006}. The third dataset which is been used in this research was collected from the City of Montreal and considered due to a large database of cast iron water pipes \cite{Zangenemadar2016}. The last dataset was from Denver water’s distribution system, which selected due to the mixture of old and new materials ranging from “pit” cast iron installed in the late 1800s to recently installed polyvinyl chloride (PVC). There was a total of 4,112 different pipes constituting the 296 miles (476.36 km) of water mains with a total of 5,610 break records \cite{Rogers2011}. All above datasets include 7 variables (pipe age, pipe installation year, pipe length, pipe material, pipe diameter, wall thickness loss, and the number of breaks) and one target variable which is the Remaining Useful Life (RUL) of the pipes. These variables were selected for inclusion in the model due to their accuracy and availability for the entire dataset.

\begin{figure}
\centering
\label{fig:fig1}
\includegraphics[width=6in]{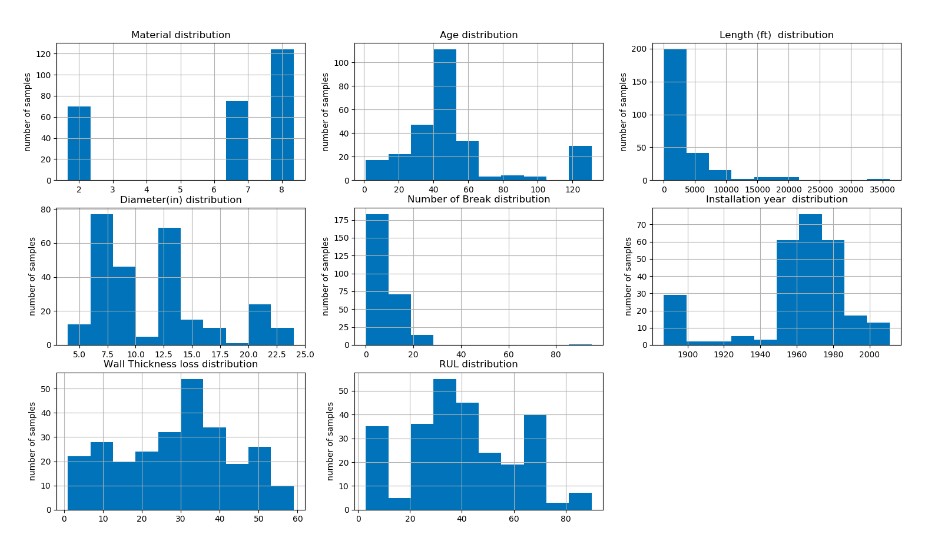}
\caption{Distribution of inputs and target variables}
\end{figure}

An initial Exploratory Data Analysis (EDA) has been performed on the dataset to find out missing data, the correlation between variables, and their distribution. Moreover, since there were a few missing data for some of the variables in the entire dataset, we have cleaned (removed) the missing values. In addition, the categorical variables have been converted into numeric data since dealing with numeric data is often easier than categorical data \cite{coso2015}.  Figure \ref{fig:fig1} illustrates the distribution of inputs and target variables of the dataset.

As can be seen from the above figure, most of the pipes were cast iron pipes constituting about 46\% of the sample with asbestos cement pipes is the second-highest number followed by ductile iron and steel pipes. The age of pipes in this study ranges between 1 and 130 years. Most of the water pipes are in the 4 to 6 in. (100 mm to 150 mm) category and pipe diameter ranges are from 4 in. to 24 in. (100 to 635 mm). The pipe installed from 1887 to 2011. Most of the pipes installed from 1960 to 1969. Majority of pipes are between 303 feet (92.35 m) and 800 feet (243.84 m). that most of the pipes break from 0 to 2 and 8 to 10 during their life that most of the water pipes have wall thickness loss around 30-40\%. The calculation of remaining useful requires data on the installation year, pipe material, and breakage history. The installation year of the pipe determines the age of the pipe. Pipe material regulates the manufacturer’s recommended service life, given as a range, which does not consider other factors such as pipe diameter \cite{Nemeth2016}. According to Devera (2013), RUL is the difference between the pipe’s age and Anticipated Service Life. Devera (2013) calculated anticipated service life (ASL) as the mean of the manufacture’s service life due to the uncertainty in the service life of a pipe. In this research, the remaining useful life is predicted based on the effects of each independent variables (age, diameter, installation year, material, number of breaks, length and wall thickness loss), Incorporating additional factors attempts to minimize the uncertainty and variation in a pipe’s service life based on additional operating conditions because the data required for this model is usually available at municipalities. The histogram shows most of the remaining useful life varies from 35 to 45 years.

Data analysis involved all collected data as means of defining initial factors affecting remaining useful life of water pipe. Furthermore, the analysis was used as means of revealing data inconsistencies and errors. Minimum, maximum, mean, mode, standard deviation, and correlation values were developed for all factors (Table 1). The correlation of an input provides an indication of whether an input will correctly, or acceptably, train with a neural network. Multiple regressions and one-way analysis of variance were generated to check the correlation between each variable (input) with remaining useful life (target). In addition, analysis of variance and t-test are generated to determine the statistical significance of the other input variables \cite{Montgomery2010}. When a model passes both ANOVA and t-test, it is statistically significant (P-value <0.05), which means that the dependent variable (response) and the independent variables (predictors) have a significant relationship. The ANOVA test results showed the significance of input variables (the p-value is p < .05).

A regression model tries to find the best fit between the actual data and the predicted value of the model \cite{Faraway2016}. The performance of the model is assessed by calculating coefficient of determination ($R^2$) or fit index.

The input with a good correlation (value close to either 1 or - 1) will usually be more significant than an input with a poor correlation (value close to 0). Pipe materials are classified based on Table 2 \cite{Zangenemadar2016}. Fares, 2010; developed a hierarchical fuzzy expert system (HFES) to estimate the risk of water main failure. The author developed a pipe material factor performance and attributed the impact value based on the risk of failure of different pipe materials. The author applied that model to different case studies to verify the model \cite{Fares2010}.

\begin{table}
\caption{Statistics of Water Pipe Database Used in this paper.}
\centering
\begin{tabular}{lccccc}
\toprule
\textbf{Features} & \textbf{Min} & \textbf{Max}  &  \textbf{Mean} & \textbf{Std} &   \textbf{Mode} \\
\midrule
Age (Years) & 1 & 131 & 49.78 & 30.31 & 43 \\
Diam.(in.) & 4  & 24 & 10.66 & 5.13 & 6 \\
Len. (ft) & 20.5 & 36,161.4 & 2,870.51 & 5,008.58 & 5,280 \\
Mat. &  1.67 &  8.35 & 6.146 &  2.75 & 8.35 \\ 
No. of Breaks & 0 & 95 & 5.09 &  7.74 & 6 \\
Ins. Year & 1,887 & 2011 & 1,961.15 & 28.78 & 1,969 \\ 
Wall Thick L. & 1 &  59 & 29.64 & 14.81 & 33 \\
RUL (Years) & 3 & 90 & 40.65 & 20.46 & 36 \\
\bottomrule
\end{tabular}
\end{table}

Based on Table 1, if we assumed the average age of pipes is 50 years, and standard deviation is approximately 30, approximately 68\% of the pipes have the age between 20 and 80 years because the amount of Z-factor is between -1 and 1 and area of -1<= Z <=1 is 68\%. The Z-factor is a measure of how many standard deviations below or above the population mean a raw score is. A Z-factor is also known as a standard score and it can be placed on a normal distribution curve. Z-factor range from -3 standard deviations (which would fall to the far left of the normal distribution curve) up to +3 standard deviations (which would fall to the far right of the normal distribution curve) \cite{Montgomery2010}.

\begin{table}
\caption{Classification of Pipe Material \cite{Zangenemadar2016}.}
\centering
\begin{tabular}{lclclc}
\toprule
\textbf{Pipe Material} & \textbf{Chance of Deterioration} &  \textbf{EA} \\
\midrule

Polyethylene & Extremely low & 0.42 \\
Ductile iron & Very low & 1.67 \\
PVC & Very low & 1.67 \\
Steel & Very low & 1.67 \\
Concrete & Medium & 5.01 \\
Asbestos & High & 6.68 \\
Cast iron & Very high & 8.35 \\
\bottomrule
\end{tabular}
\end{table}

\begin{figure}
\centering
\label{fig:fig2}
\includegraphics[width=4.5in]{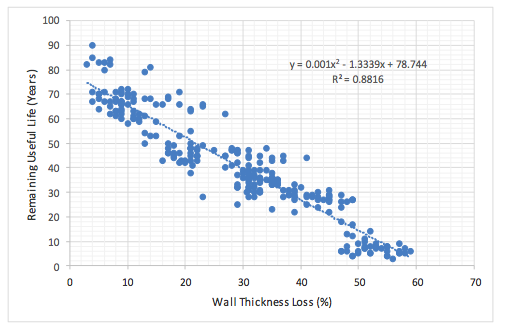}
\caption{Relationship between Remaining Useful Life and Wall Thickness Loss}
\end{figure}

Figure \ref{fig:fig2} illustrates the relationship between wall thickness loss and remaining useful life in the dataset. Remaining useful life has a lower value with the increase of wall thickness loss. This relationship is a polynomial regression. However the relationship between wall thickness loss and remaining useful life is linear here, but it does not indicate the relationship is always linear. The wall thickness loss depends on pipe material and may vary along the length of the pipe segment depending on the rate of corrosion along the pipe segment \cite{Yasseri2016}.

Figure 3 illustrates the relationship between the water ages and the remaining useful life. As age of the pipes increases remaining useful life decreases. Pipes with an age of 100 years or more have lowest remaining useful life with a quadratic regression model of 82\%. As can be inferred from Figure 3, the relationship between those variables is not linear.  Therefore, there is a need to implement the other computational models such as ANN and ANFIS to create an accurate model.

\begin{figure}
\centering
\includegraphics[width=4.5in]{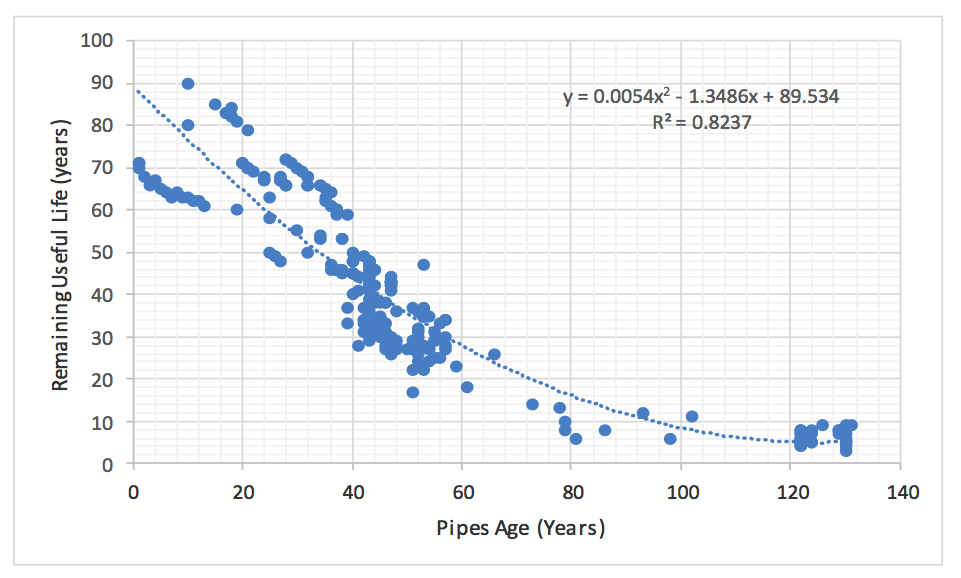}
\caption{Relationship between Remaining Useful Life and Age of Pipes}
\end{figure}

\section{Methods}

\subsection{Artificial Neural Networks in Pipeline Prediction approach}

\begin{figure}
\centering
\label{fig:fig4}
\includegraphics[width=4.5in]{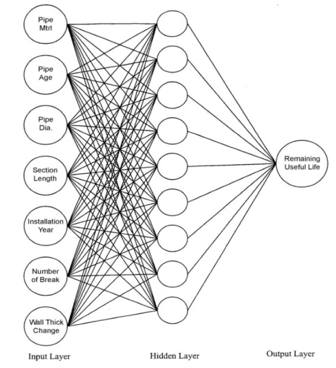}
\caption{Frameworks of ANFIS (MATLAB R2017)}
\end{figure}

Over the past decades, Artificial Neural Networks (ANNs) have been recognized as an alternative to traditional statistical models. ANNs use an algorithm inspired by research into the human brain which can “learn” directly from the data. It can be defined as “highly simplified models of the human nervous system, exhibiting abilities such as learning, generalization, and abstraction” \cite{Najafi2015}. One of the advantages of a neural network model is that a well-defined mathematical process is not required for algorithmically converting the input into an output. Once trained, a neural network can perform classification, clustering and forecasting tasks. An ANN model was chosen for this research because of its ability to cover nonlinear and compound behavior of water networks. Furthermore, it covers many variables that increase the system’s performance reliability \cite{Najafi2015}. In this research, eight ANN models developed with one hidden layer that is different in two aspects: the number of neurons in the hidden layer and the random groups of data sets. The ANN1, ANN2 ... and ANN8 models have 3, 4, 5, 6, 7, and 10 neurons in their hidden layers, respectively (Figure \ref{fig:fig4}). There is a close relationship between age, length, material, wall thickness loss, and remaining useful life. Thus, pipe material, wall thickness loss, length, diameter, and age are selected from the dataset as the input variables for ANN models. Due to the difficulty of accessing to wall thickness loss for the majority of the pipelines, this factor has been removed from five ANN models. The number of hidden layers was determined through trial runs of the model. The dataset is split randomly into training (75\%), validation (10\%) and testing (15\%). For each ANN model, trials were performed to reach the lowest error. The performance of the models was assessed based on $R^2$, mean absolute error (MAE), relative absolute error (RAE), root-relative square error (RRSE), and mean absolute percentage error (MAPE) according to Zangenemadar and Moslehi, 2016 \cite{Zangenemadar2016}.

\subsection{ANFIS Approach}

Adaptive Neural Fuzzy Inference System (ANFIS) creates a fuzzy inference system (FIS) whose membership function parameters are (adjusted) using either a back-propagation algorithm or in combination with a least squares type of method. This allows fuzzy systems to learn from the data they are modeling \cite{Suparta2016}. ANFIS works when the input that comprises the actual value is transformed into fuzzy values using the fuzzification process through its membership function, where the fuzzy value has a range between 0 and 1 \cite{Suparta2016}. ANFIS techniques provide a method for the fuzzy modeling procedure to learn information about a dataset to calculate the membership function parameters that allow the related fuzzy inference system to track the given input/output data \cite{Ghiasi2016}.

ANFIS is employed to model the relationship between the input variables. The same set of variables as applied with the ANN are considered in the fuzzy inference system. The complete ANFIS consists of five layers, the fuzzy layer, production layer, normalization layer, de-fuzzy layer, and total output layer. Each layer includes several nodes, which are defined by the node function.

\section{ANN Results}
Table 3 presents the calculated the value of MAE, PRSE, MAPE and RAE for training, testing and validating phases for the model one.

\begin{table}
\caption{ANN1 results}
\centering
\begin{tabular}{lclclc}
\toprule
\textbf {Phases} & \textbf{MAE} & \textbf{RRSE} &  \textbf{MAPE}  &  \textbf{RAE} \\
\midrule

Training  & 0.17 &  0.012 & 1.076 & 0.007 \\
Validation  & 1.304 & 0.001 & 8.047 & 0 \\
Testing & 0.88 & 0.001 & 5.431 & 0.007 \\

\bottomrule
\end{tabular}
\end{table}

\begin{figure}
\centering
\label{fig:fig5}
\includegraphics[width=4.5in]{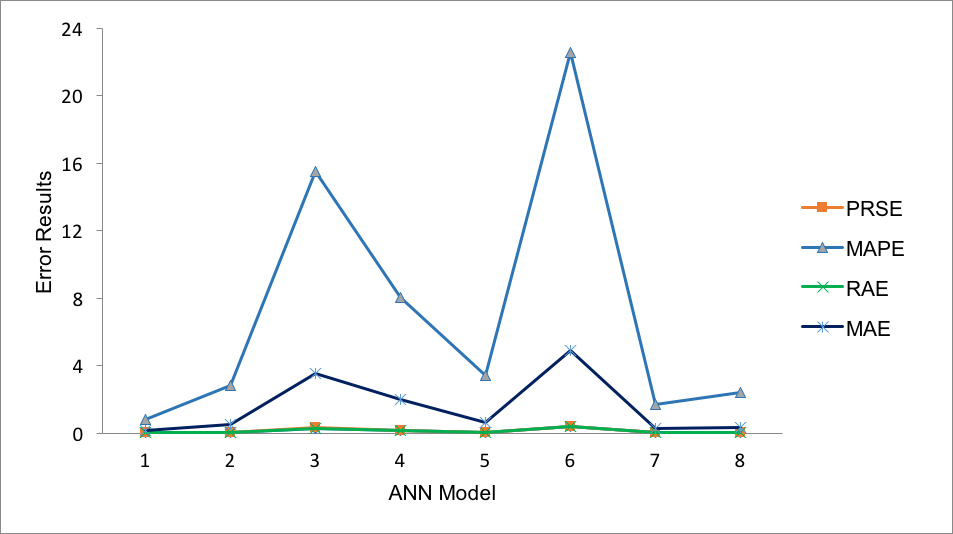}
\caption{Error Results for all the Samples}
\end{figure}

Figure \ref{fig:fig5} presents the error results for the total samples of ANN models. The X-axis presents the eight ANN models and the Y-axis depicts all error results (RRSE, MAPE, RAE, and MAE). The results demonstrates that the PRSE values are approximately equal in all ANN models in the three phases except in ANN6. The PRSE values are higher in the ANN6 model in training, validation and testing, which proves the previous assumption that this model is the least accurate one. The MAPE values are higher in validation rather than testing and training; however, the differences are more than 10\%. Because MAE, RAE, RRSE, and MAPE are the least for the ANN1 and ANN7 model, it appears to be the most precise models. Moreover, the MAPE value is less than 10, which categorizes this forecasting as a high-accuracy prediction.

\begin{figure}
\centering
\label{fig:fig6}
\includegraphics[width=4.5in]{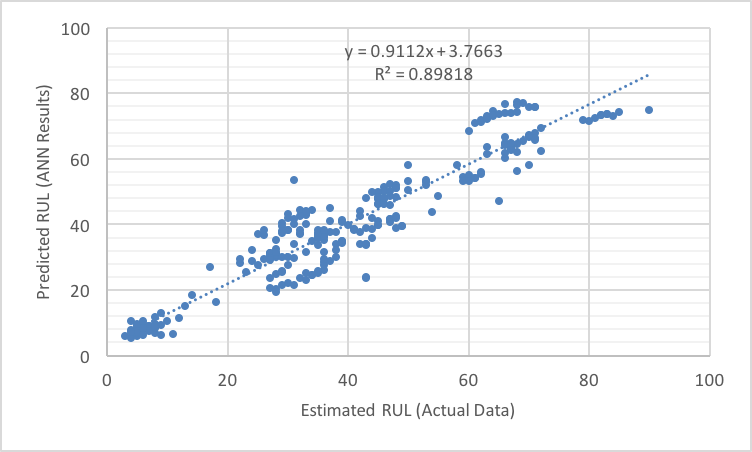}
\caption{Predicted results versus estimated results.}
\end{figure}

The comparison of predicted results and estimated results are shown in Figure \ref{fig:fig6} for the best ANN model. Estimated RUL is calculated based on actual data and predicted RUL is based on ANN results for best model. Most of the results fall in the near area of y=0.9112x+3.7663. The coefficient of determination is 89\%, which indicates that the proposed models have predicted the remaining useful life of the pipes, and is reliable for further analysis of the network. The model’s precision can be increased with adding more input variables that can affect the water conditions.

\section{ANFIS Results}
\begin{figure}
\centering
\label{fig:fig7}
\includegraphics[width=5in]{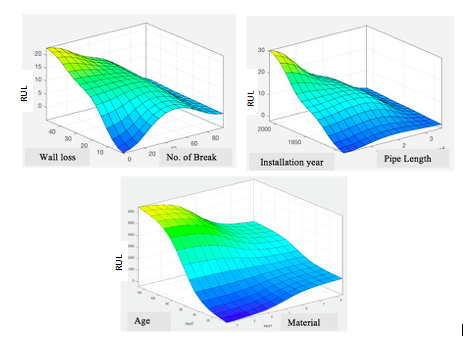}
\caption{Contour Surface for Relationship between Input variables and Output}
\end{figure}

The training data is imported into Fuzzy Logic Toolbox, and a membership function is selected. The chosen membership functions is Gaussian function. After the model is trained using the hybrid-learning rule, the results output by different membership functions were tested against the verification data. In addition, the precision of each membership function is determined using the root-mean-square-error (RMSE). Figure \ref{fig:fig7} shows the correlation of the input variables with RUL.  The slopes of wall thickness, age and installation year are higher than the other variables. Therefore, those variables have the most impact on the remaining useful life.

\section{Discussion of Results}
The results of this paper show that neural network and ANFIS were adept in capturing the relationships for prediction of remaining useful life. According to neural network results, age, and wall thickness loss were most significant parameters in RUL, while based on ANFIS, age, wall thickness loss, installation year are significant. From statistical analysis, age and wall thickness loss were the most important variables. Therefore, it is concluded that age and wall thickness loss have the most significant relationships with remaining useful life \cite{tavakoli2018}.

\begin{figure}
\centering
\label{fig:fig8}
\includegraphics[width=4.5in]{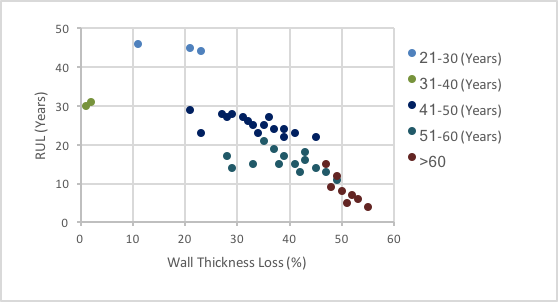}
\caption{Remaining Useful Life Prediction for Cast Iron Pipes.}
\end{figure}

Figure \ref{fig:fig8} illustrates the relationship between wall thickness loss and remaining useful life in dataset in different ages for cast iron pipes. The results show that indeed with increasing wall thickness loss remaining useful life decreases and pipes in old ages have a high renege of wall thickness loss compared to the pipes in young ages. The results show with increasing 8\% of wall thickness loss for pipes greater than 60 years old, the remaining useful life decreases 70\% approximately. Similarly, with increasing 20\% of wall thickness loss for pipes between 50 to 60 years old, the remaining useful life decreases 25\% approximately.

\begin{figure}
\centering
\label{fig:fig9}
\includegraphics[width=4.5in]{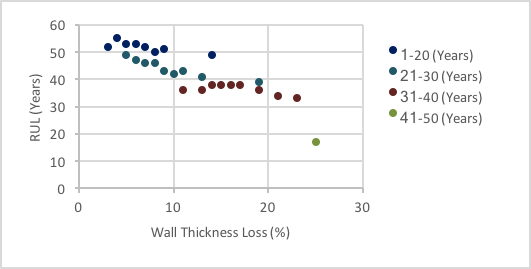}
\caption{Remaining Useful Life Prediction for Ductile Iron Pipes.}
\end{figure}

Figure \ref{fig:fig9} illustrates the relationship between wall thickness loss and remaining useful life in dataset in different ages for ductile iron pipes. The results show that indeed pipes in old ages have a high renege of wall thickness loss compared to the pipes in young ages. The results show with increasing 12\% of wall thickness loss for pipes between 31 to 40 years old, the remaining useful life decreases 10\% approximately.  Similarly, with increasing 14\% of wall thickness loss for pipes between 21-30 years, the remaining useful life decreases 20\% approximately.

\begin{figure}
\centering
\label{fig:fix10}
\includegraphics[width=4.5in]{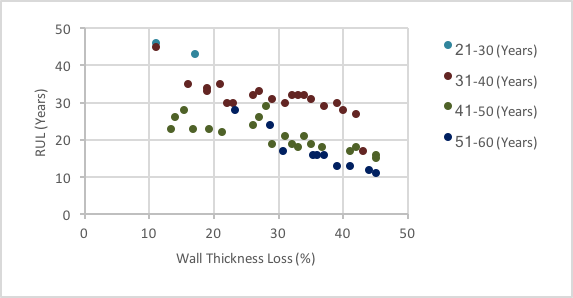}
\caption{Remaining Useful Life Prediction for Asbestos Cement Pipes.}
\end{figure}

Figure \ref{fig:fix10} illustrates the relationship between wall thickness loss and remaining useful life in dataset in different ages for asbestos cement pipes. The results show that indeed pipes in old ages have a high renege of wall thickness loss compared to the pipes in young ages. The results show with increasing 20\% of wall thickness loss for pipes between 51 to 60 years old, the remaining useful life decreases 60\% approximately.  Similarly, with increasing 30\% of wall thickness loss for pipes between 41-50 years old, the remaining useful life decreases 40\% approximately.

\begin{figure}
\centering\  
\label{fig:fig11}
\includegraphics[width=4.5in]{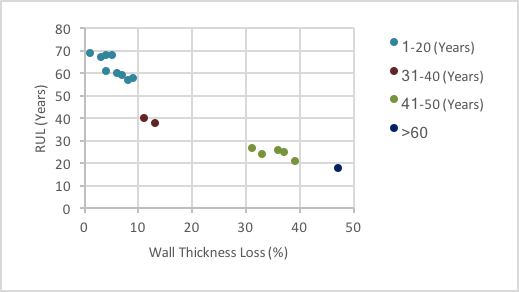}
\caption{Remaining Useful Life Prediction for Steel Pipes.}
\end{figure}

 Figure \ref{fig:fig11} illustrates the relationship between wall thickness loss and remaining useful life in dataset in different ages for steel pipes. The results show that indeed pipes in old ages have a high renege of wall thickness loss compared to the pipes in young ages. The results show with  increasing 8\% of wall thickness loss for pipes between 1 to 20 years old, the remaining useful life decreases 23\% approximately.  Similarly, with increasing 8\% of wall thickness loss for pipes between 41-50 years old, the remaining useful life decreases 20\% approximately.

\section{Validation and Contribution to the Body of Knowledge}
The deterioration models are determined with the most significant variables (age and wall thickness loss). Variables are added into the non-linear multi-variable regression ($X_1$: age, $X_2$: wall thickness loss and Y: remaining useful life). The regression models are selected based on high correlation with variables started from degree one and repeated the process with degree two and three to find the best correlation and high value of the coefficient of determination. Table 4 presents deterioration models for different types of water pipes.

\begin{table}
\caption{Deterioration Models for Different Pipes Material in the Dataset.}
\centering
\begin{tabular}{lccc}
\toprule
\textbf{Pipe Mat.} & \textbf{Non-linear Regression Models} &  \textbf{$R^2$} \\
\midrule

CI & $Y= -0.342A^2 + 0.0548W + 48.163$ &  0.78 \\
DI &  $Y= 0.004A^3 -0.025W^2 + 0.11AW + 51$ & 0.74 \\
Ac &  $Y= 0.0038A^2 -0.49W + 195.92$ & 0.80 \\
Steel & $Y= 0.005A^3 -0.012W^2 -0.989AW - 0.012$ &  0.73 \\
\bottomrule
\end{tabular}
\caption*{Where "A" is age of pipes, "W" is wall thickness loss , and "Y" is RUL.}
\end{table}

The major contributions of this paper are:
\begin{itemize}
\item This paper predicts remaining useful life of water pipes using combination of Artificial Neural Networks (ANNs) and ANFIS.
\item Above models have not been used in previous literature for determination of remaining useful life of water pipes.
\end{itemize}

\section{Conclusions and Limitations of this research}

In this paper, we have analyzed the remaining useful life of water pipes. It is concluded that the applications of neural networks and Adaptive Neural Fuzzy Inference System (ANFIS) to solve the problem of remaining useful life prediction of water pipes is feasible and the precision of the model depends on obtaining a larger and more comprehensive sample pipe dataset. Moreover, pipe age and wall thickness loss are the most significant parameters to predict the remaining useful life of the water pipes. Additionally, ductile iron and steel pipes have more remaining useful life compared to cast iron and asbestos-cement pipes. The availability of fewer numbers of deterioration parameters and limited data availability posed the primary disadvantage to effective neural network training and caused the main limitation to this paper. Environmental parameters affecting the pipe, such as overburden pressure, soil type and properties, underground water-table location and other factors identified in the literature were omitted due to lack of monitoring of the data. Review of literature showed these parameters are suitable measures for prediction of remaining useful life.

Since the developed model does not include several parameters thought to be important to water deterioration, the model developed in this paper is not complete. While it determines the utility of using Artificial Neural Networks and ANFIS models for predicting water condition, further work for data collection and model development is required to confirm that the model is more precise and reliable for future applications. The availability of detailed soils parameters, water-table location, fluctuation, joint condition, leakage history, water pressure, installation depth, temperature, and water corrosive conditions would be resources to model the deterioration of water and precisely predict the remaining useful life. Moreover, tho model does not apply to the new pipes with the age of zero and no wall thickness loss. The neural network and Adaptive Neural Fuzzy Inference System (ANFIS) for water remaining useful life prediction models, can then be combined with an inclusive infrastructure asset management system to aid the municipal agencies in better planning and spending of their limited available budget.

\bibliographystyle{unsrt}  

\bibliography{references} is enabled.

\end{document}